\pdfoutput=1

\documentclass[11pt]{article}

\usepackage{EMNLP2022}

\usepackage{times}
\usepackage{latexsym}
\usepackage{graphicx}

\usepackage[T1]{fontenc}

\usepackage[utf8]{inputenc}

\usepackage{microtype}

\usepackage{inconsolata}

\title{End-to-End Speech Translation of Arabic to English Broadcast News}

\author{  {Fethi Bougares}\\
	Le Mans Université - France\\
	
  \texttt{fethi.bougares@univ-lemans.fr} \\\And
Salim Jouili \\
  Elyadata - Tunisia \\
  \texttt{salim.jouili@elyadata.com} \\
  }

\begin{document}
\maketitle
\begin{abstract}
Speech translation (ST) is the task of directly translating acoustic speech signals in a source language into text in a foreign language. 
ST task has been addressed, for a long time, using a pipeline approach with two modules : first an Automatic Speech Recognition (ASR) in the source language followed by a text-to-text Machine translation (MT).
In the past few years, we have seen a paradigm shift towards the end-to-end approaches using sequence-to-sequence deep neural network models. 
This paper presents our efforts towards the development of the first Broadcast News end-to-end Arabic to English speech translation system. 
Starting from independent ASR and MT LDC releases, we were able to identify about 92 hours of Arabic audio recordings for which the manual transcription was also translated into English at the segment level.
These data was used to train and compare pipeline and end-to-end speech translation systems under multiple scenarios including transfer learning and data augmentation techniques. 
\end{abstract}

\section{Introduction}

End-to-end approach to speech translation is gradually replacing the cascaded approach which consists of transcribing the speech inputs with an ASR system, and translating the obtained transcription using a text-to-text MT system. For instance, and for the first time, the winning system in the IWSLT 2020 TED English-to-German speech translation shared task was an end-to-end system \cite{ansari-etal-2020-findings}. Despite this positive result, the end-to-end approach is used on a limited scale due to the lack of labeled data. Indeed, data scarcity is today the major blocker for the widespread adoption of the end-to-end models. Taking this into consideration, recent works have focused on developing speech translation corpora. Joint efforts in this direction have allowed us to
collect a significant quantity and good quality of speech translation corpora. Not surprisingly, speech translation corpus development has started for well-resourced languages including English and some  European languages. In \cite{1802-03142}, the 236 hours English$\rightarrow$French ST Augmented LibriSpeech were released. Shortly after, \cite{di-gangi-etal-2019-must} released the MUST-C corpus including few hundreds (385 to 504 hours) of parallel ST data of TED talks translations from English to eight European languages. At the same time, the Europarl-ST \cite{abs-1911-03167} was released with translations between 6 European languages, with a total of 30 translation directions. 
While all the previous resource development effort has focused on well-resourced languages, the most recent published corpora CoVoST \cite{wang2020covost} and CoVoST2 \cite{wang2020covost2}. These latter works released a large-scale Multilingual Speech-To-Text Translation Corpus covering translations from 21 languages to English and from English to 15 languages. Although the Arabic-English is one of the language pairs covered by the CoVoST2 corpus, the authors consider it as a low-resource pair. In fact, CoVoST2 corpus contained only 6 hours of prepared speech uniformly splitted between train, dev and test sets. In this paper, we conduct a series of experiments to present the first results of Arabic to English End-to-End Broadcast News Speech translation.

This paper is organized as follows: section 2 presents Arabic-to-English speech translation related works. Section 3 gives details about the source of our training data and the method we have used to extract theses data. In section 4, we present our experimental setup as well as the used toolkits to train ou models. Sections 5 and 6 provides details about the pipeline and end-to-end speech translation systems, respectively. Section  7, gives a brief discussion and results analysis. Finally, section 8 concludes the findings of this study and discusses future work.


\section{Related works}
\label{sec:format}

Arabic-English (AR-EN) is one of the most studied language pair in the context of Speech Recognition and Machine Translation. For instance, this language pair was integrated in several evaluation campaigns and projects including the International Workshop on Spoken Language Translation (IWSLT) and DARPA's Babylon project. These earlier projects have built on the traditional pipelined architecture integrating speech recognition system in the source language followed by machine translation from the transcript to the target language. In IWSLT, the speech translation task was introduced for the first time in 2006. The IWSLT06 \cite{iwsltPaul06} translation campaign was carried using either the manual or the automatic transcription of speech input in the travel domain. This translation task was renewed for several years using always the pipeline approach.\\

Pipeline architecture was also used by BBN in the context Babylon project \cite{Stallard11}. They developed the TransTalk system including a pipeline of ASR and MT systems in both directions (AR$\leftrightarrow$EN). More recently, but still with the same approach, QCRI presented their live  Arabic-to-English speech translation system in \cite{dalvi-etal-2017-qcri}. The system is a pipeline of a Kaldi-based Speech Recognition followed by a Phrase-based/Neural MT system. Recently, there has been a shift to the most recent end-to-end approach without the intermediate step of transcribing the source language. Indeed, IWSLT 2018 was the first time where  organizers drooped the ASR task and participants needed to develop an end-to-end speech translation systems. End-to-end speech translation has shown its effectiveness for multiple languages and in multiple scenarios. It becomes now a well-established task in IWSLT evaluation campaign were multiple shared taks are proposed to assess Spoken Language Translation (SLT) systems for many language pair in several settings. Despite the great interest being shown to the end-to-end approach for speech translation task, we were able to identify only one recent work by \cite{wang2020covost2} including Arabic-English language pair limited to a corpus of 6 hours. We are also aware of the IWSLT 2022 Dialectal Speech Translation\footnote{https://iwslt.org/2022/dialect} task which, unlike this work, focuses on Tunisian-to-English speech translation \cite{zanon-boito-etal-2022-trac,yan-etal-2022-cmus,yang-etal-2022-jhu}.


\section{Training Data}
\label{sec:pagestyle}

Whatever the chosen architecture for Speech translation system (pipeline or end-to-end), it requires a large amount of manually annotated training data that might be hard to obtain for multiple language pairs. 
For the Arabic-English language pair, a large amount of training data for ASR and MT was manually annotated in the framework of the DARPA's Global Autonomous Language Exploitation (GALE) project \cite{4430115}. 
This huge amount of work was done for the purpose of making foreign languages speech and text accessible to English-only speakers through the development of automatic speech recognition and machine translation systems.\\

In this respect, Arabic broadcast news and conversation speech were collected from multiple sources, then annotated under the supervision of Linguistic Data Consortium (LDC). 
Audio corpora and their  transcripts are separately released in three phases : GALE Phase 2, 3 and 4. In addition to the speech audio corpora and transcripts released to train Arabic ASR systems, LDC also made available multiple Arabic to English parallel corpora.The latter are intended to be used for training and evaluating Arabic to English MT systems. They have been developed by manually translating from a number of different sources including Arabic news-wire, discussion groups and broadcast news and conversation.\\

Upon closer inspection of these parallel corpora, we have found that part of the broadcast news and conversation parallel corpora were built by translating the manual transcripts released for the ASR task.
Following the discovery we dug deep in the GALE speech recognition and machine translation LDC related releases and, as illustrated in Figure 1, we parsed all GALE speech recognition and machine 
translation corpora in order to extract a 3-way parallel corpus consisting of Arabic audio along with their Arabic transcriptions and English translation. As shown in Figure 1, for each transcribed audio file part of the GALE corpus, we extract only segments for which we were able to find an exact match between the manual transcription, from ASR training data, and the source side of parallel corpora. 
Table \ref{tab:Extractdata} shows the amount of speech audio for which we were able to find the corresponding translation in the LDC MT related releases. We report, for each phase: 2, 3 and 4, the original size of the 
speech corpus in hours and the extracted subset for which the English translation had been found.\\

\begin{figure*}[h]
\centering
\includegraphics[scale=0.45]{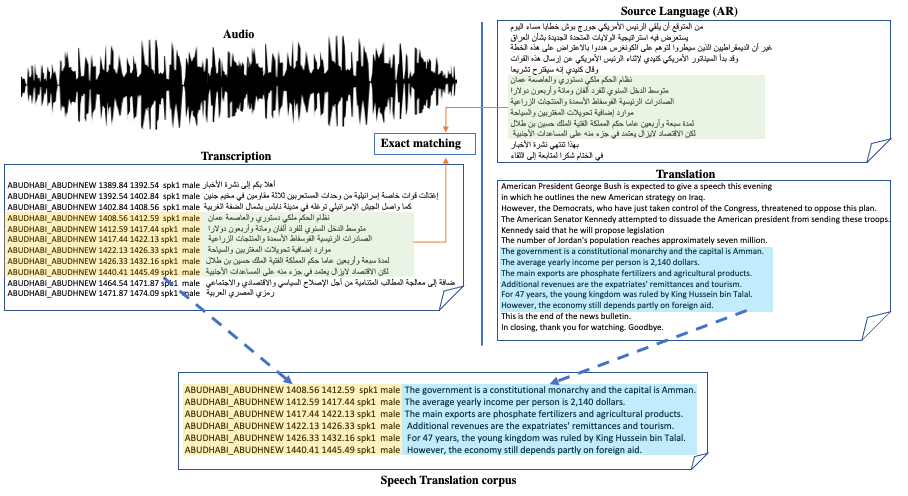}
\caption{Extraction Arabic to English speech translation corpus from LDC ASR and MT independent releases.}
\label{CorpusExtract}
\end{figure*}

\begin{table}[h!]
\centering
\begin{tabular}{lrr}
\hline
\textbf{GALE Phase}       & \multicolumn{1}{c}{\textbf{Hours}} & \textbf{\#Segments}  \\
\hline \hline
\textit{Phase 2} & 436h 11 min & 190.510 \\
\textit{Phase 2 ST. } &  59h 12 min & 24.519\\ \hline
\textit{Phase 3} & 419h 03 min & 195.143 \\
\textit{Phase 3 ST.} & 28h  50 min & 13.261  \\ \hline
\textit{Phase 4} &  96h 18 min & 54.787 \\
\textit{Phase 4 ST.} & 4.0h 08 min & 1.855 \\
\hline
\textit{Phase 2/3/4} & 951h 32 min  & 440.440 \\
\hline
\textit{Extracted ST.} & 92h 10 min & 39.635 \\
\hline
\end{tabular}
\caption{Statistics of the original GALE Arabic to English Speech Transcription corpus and the extracted subsets for which translations are available.}
\label{tab:Extractdata}
\end{table}

All the extracted segments were afterwards aligned using timestamps information from ASR transcript and translation from MT target side. As table \ref{tab:Extractdata} shows, an overall Arabic to English speech translation corpus of around 92 hours was extracted. This corpus was then cleaned out by removing all the back-channel and incomplete speech segments. The final corpus is then splitted into training, development and test sets.
Development and test contain segments from randomly selected broadcast audio programs. Their size is respectively 1188 and 987 segments. Development set contain broadcast News and Conversation recordings from Abu Dhabi TV, Al Alam News Channel, based in Iran and Al Arabiya. Test set is made up of broadcast News and Conversation recordings from Abu Dhabi TV, Aljazeera, Al Arabiya and Syria TV. The remaining material was used as training data for ASR, MT and ST systems.\\

\begin{table}[h!]
\centering
\begin{tabular}{lccc}
\hline 
& \textbf{Train} & \textbf{dev.} & \textbf{test} \\
\hline \hline
\textit{Hours} & 83h54 & 3h05 & 2h38\\
\textit{Sentences} & 32.099 & 1188 & 987 \\
\textit{\#AR words}& 606.465 & 22.537 & 18.598  \\ 
\textit{\#EN words}& 945.801 & 35.180 & 27.880\\ 
\hline
\end{tabular}
\caption{Statistics and splits of the extracted Arabic to English Speech Translation corpus extracted from LDC ASR and MT independent releases.}
\label{tab:data}
\end{table}

Table \ref{tab:data} gives a detailed statistics of the extracted Arabic to English Speech Translation corpus, including speech duration as well as token counts for both transcripts and translations.


\section{Experimental Setup}

All our experiments are built using open source toolkits with the following settings:
ASR models were built using the End-to-End Speech Processing Toolkit ESPnet \cite{abs-1804-00015}.  We trained an attention-based encoder-decoder architecture with an encoder of 4 VGGBLSTM layers including 1024 cells in each layer. The second and third bottom BLSTM layers of the encoder reduced the utterance length by a factor of two. We used a decoder of one 1024-dimensional BLSTM layer. For both ASR and ST speech utterance, we extracted 40 Melfilterbank energy features with a step size of 10ms and a window size of 25ms. The extracted features, we applied mean and variance normalization. MT models were built using the FAIRSEQ package \cite{ott2019fairseq}. We trained end-to-end word and \textit{bpe-based} translation systems using the \emph{"lstm\_luong\_wmt\_en\_de"} model template. This template is a standard LSTM Encoder-Decoder architecture composed of 4 stacked BLSTM layers, each with 1000 cells, and 1000-dimensional embeddings  \cite{luong-etal-2015-effective}. 
Translation tasks (AST and MT) evaluation was carried out using case-sensitive BLEU metric (Papineni et al., 2002). Scores are calculated using one human reference with Moses’mteval-v14.pl script \footnote{\url{https://github.com/moses-smt/mosesdecoder/blob/master/scripts/generic}} applied to de-tokenized and punctuated translation output. As for ASR,  systems were evaluated using Word Error Rate (WER).


\section{Pipeline Speech Translation}
\label{lowres}

In this section, we evaluate the pipeline approach for speech translation in two different scenarios, plausible for many language pairs, depending on the amount and the type of training data used for the development of the Speech Translation task.

\begin{enumerate}
    \item \textbf{Constrained Scenario } : Under this scenario we have access to a 3-way limited training data. This data includes speech audio files in source language their transcriptions in the source language and translations to the target language.
    
   \item \textbf{Unconstrained Scenario} : In addition to resources from the constrained scenario, we have access to a large ASR and MT-specific resources.\\
\end{enumerate}

As to the first scenario of the pipeline approach we only used the 3-way parallel data reported in table \ref{tab:data}. 
In this instance, an end-to-end ASR module was trained using ESPnet \cite{watanabe2018espnet} toolkit on the speech audio files from Table \ref{tab:data} and their corresponding transcripts.  
In the Unconstrained Scenario, however, ASR module was trained using the totality of the GALE Phase 2, 3 and 4 ASR data reported in Table \ref{tab:Extractdata}. 

\begin{table}[h!]
\centering
\begin{tabular}{lcc}
\hline 
\textbf{ASR System} & \textbf{dev} & \textbf{test}   \\
\hline \hline
ASR\_\emph{Const}   & 20.90 & 21.90  \\ \hline
ASR\_\emph{UnConst}  & 13.10 & 14.60  \\ \hline
\end{tabular}
\caption{ASR WER (in \%) on the dev and test sets.}
\label{tab:asr}
\end{table}

Table \ref{tab:asr} presents the results of ASR system under both constrained and unconstrained scenarios.  As shown in the Table \ref{tab:asr}, using a training set of around 84h of manually transcribed broadcast news and conversation, we obtained a WER of 20.90\% and 21.90\% on dev and test sets, respectively.  Not surprisingly, WER has been significantly improved with the use of the complete GALE training data\footnote{We have taken particular care to remove dev and test data before using GALE corpora to train the ASR system.} (row ASR\_\emph{UnConst}) to achieve \textbf{13.10\%} and \textbf{14.60\%} on dev and test sets, respectively.

As previously stated, within the pipeline ST framework, the output of the ASR module is automatically translated to the target language using the MT module. 
The MT module is also an end-to-end system trained using Fairseq toolkit \cite{ott-etal-2019-fairseq} under both constrained (MT\_\emph{Const}) and unconstrained (MT\_\emph{UnCons}) scenarios. Table \ref{tab:mt} reports the BLEU scores of the translation output by varying ASR module condition while fixing MT module constrained to speech translation data composed of the transcripts along with their corresponding English translation from Table \ref{tab:data}.\\

\begin{table}[h!]
\centering
\begin{tabular}{lcc}
\hline
\textbf{Pipeline ST System} & \textbf{dev}  & \textbf{test} \\
\hline \hline
MT\_\emph{Const\_ASR\_\emph{Const}}  & 19.03 & 15.96 \\ \hline
MT\_\emph{Const\_ASR\_UnConst}          & 20.69 & 16.58  \\ \hline
MT\_\emph{Const\_ref\_Transc}       		& 22.31 & 18.30  \\ \hline
\end{tabular}	
\caption{Case-sensitive tokenized and single-reference BLEU scores (in \%) of the pipeline speech translation system with the constrained MT module.}
\label{tab:mt}
\end{table}

The first row in table \ref{tab:mt} (MT\_\emph{Const\_ASR\_\emph{Const}}) gives the BLEU score when the MT constrain module translates the output of a constrained ASR system (row ASR\_\emph{Const} from Table \ref{tab:asr}). In this case, a BLEU score of 19.03\% and  15.96\% is respectively achieved on dev and test sets.
The second row in the same table (MT\_\emph{Const\_ASR\_UnConst}) shows the BLEU score when the ASR module is under the unconstrained condition, i.e. output from the system ASR\_\emph{UnConst} in Table \ref{tab:asr} are used as input to the MT system. As expected, when it comes to translating a higher transcription quality, the translation quality is better and the BLEU score is increased by 1.66 and 0.62 BLEU points on dev and test sets, respectively. 
The last row of table \ref{tab:mt} (MT\_\emph{Const\_ref\_Transc}) simulates the situation where we have access to a perfect transcripts in the source language. In this case, translation quality is further improved reaching 22.31 BLEU points on dev set and 18.30 points on test set.\\

In a similar vein, table \ref{tab:m1t} presents results in settings where MT module is no longer 
constrained to speech translation data. Indeed, additional Arabic to English Bilingual text from GALE LDC releases are 
used to train the unconstrained MT module \footnote{Unconstrained MT system was trained using all GALE Arabic-English Parallel Text from 2007 to 2016.}. This unconstrained MT module, was used to run several experiments using various input conditions similar to what we did within the constrained condition. 
The results of these experiments are presented in Table \ref{tab:m1t}. The first row (MT\_\emph{UnConst\_ASR\_Const}) sets out the BLEU score when the unconstrained MT module translates the output of the constrained ASR (first row in table \ref{tab:asr}). Compared to using the constrained MT system, a considerable improvement of 12.84 (from 19.03 to 31.87) and 8.26 (from 15.96 to 24.22) BLEU points is achieved on dev and test sets, respectively. \\

As we have seen above, translation quality is further improved when the input to the translation module is of a higher quality generated by the unconstrained ASR system (row MT\_\emph{UnConst\_ASR\_UnConst}). This allows to reach a dev and test BLEU scores of 36.48 and 25.80 respectively. As expected, the BLEU score is even better when it comes to translate the reference transcription (MT\_\emph{UnConst\_ref\_Transc}) as shown in the last row of Table \ref{tab:m1t}. In the latter case, we achieved a dev set BLEU score of 39.51 and a test set BLEU score of 30.60.

\begin{table}[h!]
\centering
\begin{tabular}{lcc}
\hline
\textbf{Pipeline ST System} &  \textbf{dev} & \textbf{test}   \\
\hline \hline
MT\_\emph{UnConst\_ASR\_Const}   & 31.87 & 24.22  \\ \hline
MT\_\emph{UnConst\_ASR\_UnConst} & 36.48 & 27.51  \\ \hline
MT\_\emph{UnConst\_ref\_Transc}     & 39.51 & 30.60  \\ \hline
\end{tabular}
\caption{Case-sensitive tokenized and single-reference BLEU scores (in \%) of the pipeline speech translation system with Unconstrained MT module.}
\label{tab:m1t}
\end{table}


\section{End-to-End Speech Translation}
In this section, we present and evaluate the end-to-end approach for Arabic to English speech translation task. 
The End-to-End system is built using the ESPnet toolkit \cite{abs-1804-00015}. We used 
an attention-based encoder-decoder architecture. The encoder has two VGG-like CNN blocks followed by five stacked 
1024-dimensional BLSTM layers. The decoder is composed of two 1024-dimensional LSTM layers. Each VGG block contains two 
2D-convolution layers followed by a 2D-maxpooling layer whose aim is to reduce both time and frequency dimension of 
the input speech features by a factor of 2. All our experiments are conducted using characters as target tokens.\\

Table \ref{tab:st_res} shows the performance of the end-to-end ST model with different training configurations.

\begin{table}[h!]
\centering

\begin{tabular}{llc}
\hline
\textbf{End2End ST system} & \textbf{dev}  &  \textbf{test} \\
\hline
\hline
Baseline (1)  & 2.58 & 2.23  \\ \hline
(1) + Enc. init & 12.44 & 9.57 \\ \hline
(1)  + Unsup ph234 & 23.23 & 18.97  \\ \hline
(1) + Enc. Init + Unsup ph234 & \textbf{24.95} & \textbf{19.09}  \\ \hline
\end{tabular}
\caption{Case-sensitive tokenized and single-reference BLEU score (in \%)  of the End-to-end AR$\rightarrow$EN Speech Translation system with Encoder initialization and data augmentation}
\label{tab:st_res}
\end{table}

The first row from Table \ref{tab:st_res} shows the baseline results obtained when the end-to-end model is trained under the constrained scenario, that is when the training data is restricted to the 83h54 minutes from table \ref{tab:data}. We can clearly see that the end-to-end model is not strong enough to compete with the cascaded model trained using the same amount of data. Indeed, the BLEU score of the end-to-end  system on the dev set is 2.58, compared to the 19.03 points of the pipeline model. The same goes for test set where end-to-end system BLEU score is 2.23 compared to 15.96 which is obtained with cascade translation approach.\\

From this initial baseline and with the aim of improving the end-to-end system translation quality, we employed the well established transfer learning technique \cite{abs-1809-01431} commonly referred as encoder pre-training. 
Indeed, using the ASR encoder of the Unconstrained ASR system (row ASR\_\emph{UnConst} in \ref{tab:asr}) to initialize the parameters of the ST encoder greatly improves the performance of end-to-end ST networks.
The results of the encoder pre-training are shown in the second row ( (1) + Enc. init) of table \ref{tab:st_res}.
As a result, we observed a strong effect reflected by the substantial improvement in the BLEU score: +9.86 and +7.34 BLEU score for dev and test sets, respectively.\\

Just like the transfer learning via encoder pre-training approach, data augmentation is proven to enhance end-to-end speech translation quality.
It is carried out using synthetic data which is generated by automatically translating the transcripts of an ASR corpora in the source language. Herein, we used the unconstrained NMT system (MT\_\emph{UnConst}) of table \ref{tab:m1t} in order to translate the Arabic GALE transcripts provided in table \ref{tab:Extractdata}. Incomplete and back-channel speech segments were filtered out from the generated translations. All in all, we were able to create the synthetic corpus of 795 hours of Arabic to English speech translation corpus detailed in table \ref{tab:Syntdata}.

\begin{table}[h]
\renewcommand\thetable{7}
\centering
\begin{tabular}{lcccc}
\hline
    & \textbf{Hours} & \textbf{\#Sent.} &\textbf{\#AR} & \textbf{\#EN} \\
\hline
\textit{Gale Synth.} & 795 & 314.167 & 6.1M & 9.1M \\
\hline
\end{tabular}
\caption{Statistics of the synthetic Ar-En ST corpus.}
\label{tab:Syntdata}
\end{table}

These synthetic data are thereafter used as additional data to train the end-to-end ST system. The results of this data augmentation experiment are highlighted in Table \ref{tab:st_res} (row (1) + unsup ph234). As we can see from the obtained results synthetic training data boosts up the end-to-end ST system to achieve a BLEU scores of \textbf{23.23} points and \textbf{18.97} points for dev and test sets, respectively.\\

Both encoder pre-training and data augmentation are shown to be helpful improving significantly the ST baseline. We also experimented using both methods at the same time. The last row of the same table presents the results of the end-to-end speech translation trained with data augmentation using synthetic data from Table \ref{tab:Syntdata}, and encoder hyperparameters initialization from the ASR\_\emph{UnConst} system presented in Table \ref{tab:asr}. By applying these two methods together, we were able to reach a BLEU scores of \textbf{24.95} and \textbf{19.09} points for dev and test sets, respectively. These end-to-end speech translation results are to be compared to pipeline results shown in row MT\_\emph{UnConst\_ASR\_UnConst} of  table \ref{tab:m1t}.

\begin{figure*}[h]
\centering
\includegraphics[scale=0.55]{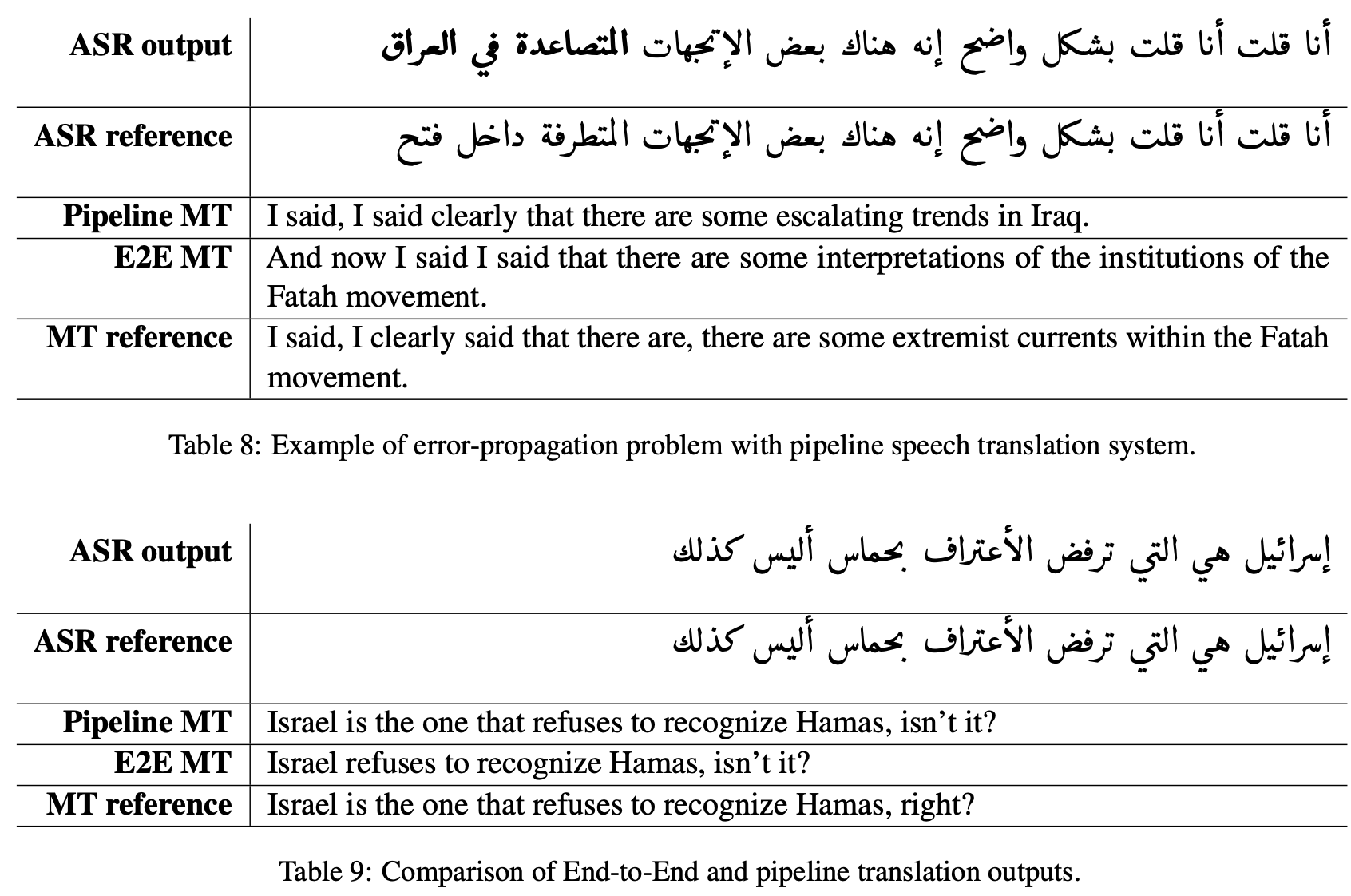}
\label{CorpusExtract}
\end{figure*}


\section{Discussion and analysis}

Despite the improvements brought by transfer learning and data augmentation technics, the best results are still obtained using cascade 
architecture. We believe that this performance gap can be partly explained by the fact that end-to-end system was trained using only a small 
amount ($\sim$ 84 hours) of real speech translation corpus.  \\

Based on the results of previous works from \cite{abs-1904-08075}, the end-to-end ST models are known  
as an effective means of circumventing the error-propagation problem faced by the conventional pipeline system. Indeed, every involved 
component in the traditional pipeline approach produces errors, which are propagated through the cascade and lead to compounding follow-up 
errors. In order to assess the ability of our end-to-end system to overcome this error-propagation pattern, we selected some translation examples 
where pipeline system fails due to this problem and we checked the translation output of the end-to-end system. 
Example from table 8 shows a translation error caused by the propagation of transcription errors occurred at the end of the segment (text in bold ASR output row). The end-to-end system, however, relies on the source speech signal and translates correctly the same part of the input.\\

In addition to this error-propagation problem we have found that end-to-end system is sometimes penalized although its translation is correct.  
Table 9 presents and example where both systems output correct translation but BLEU score is better with pipeline system. 
The thing might happen for pipeline system as well, but we believe that end-to-end systems are more affected as the translation references
are obtained by translated from a textual input, not from speech audio in the source language. This trend must be probed further in order to quantify its impact on the 
end-to-end ST system performance. We leave such investigations as future work.  

\section{Conclusions}

In this paper, we presented the first results on Arabic-to-English end-to-end Automatic Speech Translation system. Arabic-English language pairs is one of the well-studied language pair in Natural Language Processing. Therefore, large quantities of data are made available in a wide variety of domains including ASR and MT. Starting from independent LDC releases for MT and ASR systems, we  were  able  to  extract around 92 hours of speech translation corpus composed of Arabic audio and  their source transcriptions and English translation. We used this corpus to conduct speech translation experiments using a pipeline and an end-to-end Speech Translation architecture. Both methods were tested under a constrained and an unconstrained conditions. We showed that the performance gap, which is too big between the two considered approaches  under the constrained condition, can be narrowed under unconstrained condition through the use of transfer learning and data augmentation techniques. In spite of considerable improvement obtained by applying these techniques, the gap remains important and we plan to reduce it in several ways including decoder pre-training, spectrogram augmentation and Self-Supervised Learning.

\section{Acknowledgements} 
This work was funded by the ESPERANTO project. ESPERANTO project has received funding from the European Union’s Horizon 2020 research and innovation programme under the Marie Skłodowska-Curie grant agreement No 101007666.

\bibliography{anthology,custom}
\bibliographystyle{acl_natbib}

\end{document}